\pdfoutput=1
\documentclass[10pt,twocolumn,letterpaper]{article}
\pdfoutput=1
\usepackage{wacv}
\usepackage{svg}
\usepackage{times}
\usepackage{epsfig}
\usepackage{graphicx}
\usepackage{amsmath}
\usepackage{amssymb}
\usepackage{url}
\usepackage{support-caption}
\usepackage{subfig}
\usepackage{algorithm}
\usepackage{algpseudocode}

\newlength\myindent
\usepackage[pagebackref=true,breaklinks=false,bookmarks=false]{hyperref}
\hypersetup{
    colorlinks=true,
    linkcolor=magenta,
    filecolor=magenta,      
    urlcolor=magenta,
}
\wacvfinalcopy % *** Uncomment this line for the final submission
\def\wacvPaperID{536} % *** Enter the wacv Paper ID here

\ifwacvfinal\fi
\begin{document}

%%%%%%%%% TITLE
\title{A Little Fog for a Large Turn}

% Authors at the same institution
%\author{First Author \hspace{2cm} Second Author \\
% Authors at different institutions
\author{Harshitha Machiraju, Vineeth N Balasubramanian \\
Indian Institute of Technology, Hyderabad, India\\
{\tt\small \{ee14btech11011, vineethnb\}@iith.ac.in}
}

%\author{Harshitha Machiraju \\
%IIT Hyderabad\\
%{\tt\small ee14btech11011@iith.ac.in}
%\and
%Vineeth N Balasubramanian \\
%IIT Hyderabad\\
%{\tt\small vineethnb@iith.ac.in}
%}

\maketitle
\ifwacvfinal\thispagestyle{empty}\fi

%%%%%%%%% ABSTRACT
\begin{abstract}
  Small, carefully crafted perturbations called adversarial perturbations can easily fool neural networks. However, these perturbations are largely additive and not naturally found. We turn our attention to the field of Autonomous navigation wherein adverse weather conditions such as fog have a drastic effect on the predictions of these systems. These weather conditions are capable of acting like natural adversaries that can help in testing models. To this end, we introduce a general notion of adversarial perturbations, which can be created using generative models and provide a methodology inspired by Cycle-Consistent Generative Adversarial Networks to generate adversarial weather conditions for a given image. Our formulation and results show that these images provide a suitable testbed for steering models used in Autonomous navigation models. Our work also presents a more natural and general definition of Adversarial perturbations based on Perceptual Similarity. 
 \footnote{Accepted to \href{http://wacv20.wacv.net/}{WACV 2020} \tiny}
\end{abstract}

\section{Introduction}
\label{Introduction}

Autonomous navigation has occupied a central position in the efforts of computer vision researchers in recent years. Autonomous vehicles can not only aid navigation in urban areas but also provide critical support in disaster-affected areas, places with unknown topography (such as Mars), and many more. The vast potential of the applications thereof and the feasibility of the solutions in contemporary times has led to the growth of several organizations across industry, academia, and government institutions that are investing significant efforts on self-driving vehicles. Computer vision has been studied and shown to play an important role in the development of autonomous navigation technologies over the years \cite{Hebert1988}\cite{GARIBOTTO199751}\cite{JanaiGBG17}. Vision tasks such as image-level classification, object detection, semantic segmentation, as well as steering angle prediction, play critical roles in the development of autonomous vehicles. The increasing emphasis of this problem domain has also led to the creation of different vision datasets that are necessary to develop solutions across different geographies \cite{Cordts2016Cityscapes}\cite{RobotCarDatasetIJRR}\cite{varma2019idd}.

% \clearpage
\begin{figure}[ht]
\centering
\subfloat[]{
  \centering
  % include first image
  \includegraphics[width=\linewidth,height=3cm]{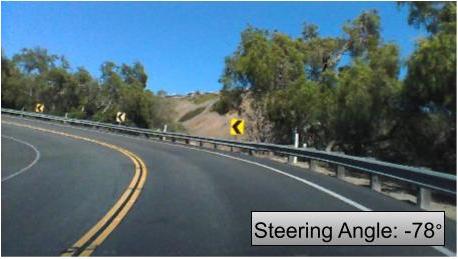}  
  \label{fig:steering_sample_orig}
}\\[-2ex] 
\subfloat[]{
  \centering
  % include second image
  \includegraphics[width=\linewidth,height=3cm]{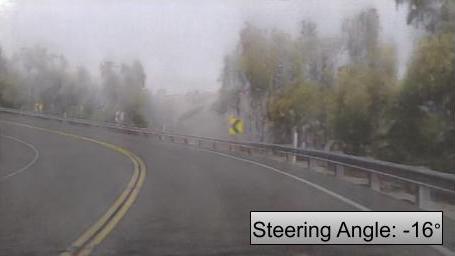}  
  \label{fig:steering_sample_fog}
}
\caption{Steering Angle(radians) deviation seen in the same scene due to Fog, for the AutoPilot model\cite{sully_chen_autopilot}. Lower image was generated by our method given the left image }
\label{fig:steering_sample}
\end{figure}

It is common knowledge now that deep neural networks have achieved state-of-the-art performance in many computer vision tasks \cite{mask_rcnn,instagram_2018}. With great leaps in performance, deep learning models have been deployed in many physical systems, and efforts have been afoot to develop robust deep neural network models for autonomous vehicles too \cite{nvidia_pilotnet}\cite{sully_chen_autopilot}\cite{comma}. However, the recent mishaps involving self-driving vehicles has necessitated the requirement for testing such deep learning models with data in various conditions. Existing efforts largely rely on copious amounts of collected data from the real world, data augmentation using simple affine transformations \cite{deeptest} or the use of data from synthetic/virtual environments \cite{synthiaCVPR2016} for various conditions. There is an impending need for methods that can provide data with a wider variety of conditions that can validate the robustness of the learned models for vision tasks in such settings, in order to save lives and property in the future. One such important dimension is the variability of a given environment under various weather conditions. Real-world studies such as \cite{weathernewsarticle2014} have shown that bad weather can alone manage to crash navigation systems. It is hence important to train deep learning models with as many weather conditions of a given environment as possible to obtain robust systems in deployment. Our efforts in this work are towards addressing this need.

From a different perspective, many recent efforts have also been made to study the robustness of deep neural network models, by showing how vulnerable they can be towards adversarial perturbations \cite{akhtar2018threat}, which involve adding a small amount of noise to the input data in order to fool the model. Attempts have been made to show that systems performing essential day-to-day tasks such as face and speech recognition \cite{face_recog_attack}\cite{carlini_speech_attack}, as well as physical systems \cite{goodfellow_physical_world} can be attacked using such adversarial perturbations. Similar efforts have also been attempt to attack autonomous navigation models such as in \cite{dawn_song_physical_attack}\cite{agv_attack_paper}. These efforts use adversarial patches \cite{adversarial_patch} physically placed in the field of view of the model to fool it. However, as shown in a very recent work \cite{hendrycks2019natural}, natural adversarial examples are sufficient to fool them, and do not need any explicit image manipulation or hacking with an intention to fool such systems. Weather-based changes in the environment fall into such a category, which we focus on in this work. For example, as shown in Figure \ref{fig:steering_sample}, when fog is added to the scene, the steering angle predicted by deep learning models deviates by a significant amount from the original, making these models vulnerable to such `weather-adversarial' data.

In this work, we bring together two perspectives: the data augmentation one and the adversarial one, to explicitly generate weather-adversarial images that can fool deep neural networks for autonomous vehicles (steering angle prediction, in particular). To the best of our knowledge, this is the first effort in this direction. Such an effort is important to provide essential data from different conditions that are likely to affect (or attack) such models. Our work can be utilized to test steering angle prediction models and their robustness against adverse weather conditions. (We focus on fog in this work, due to the availability of relevant data, but our framework is generalizable and can be easily extended to other weather conditions.) Our work also demonstrates that current steering angle models used in self-driving cars are inadequate to handle weather variations in real-world settings. The key contributions of this work are as follows: 
\begin{itemize}
\itemsep0em 
    \item We bring together data augmentation and adversarial perspectives to introduce a methodology that can generate foggy images that are intended to `fool' models for steering angle prediction in autonomous vehicles. Our methodology integrates an adversarial loss term in an unpaired image-to-image translation framework \cite{CycleGAN2017}\cite{DistanceGAN} towards the aforementioned objective. To the best of our knowledge, this is the first such effort, in particular, for autonomous navigation applications.
    \item Existing adversarial attack models are largely focused on classification tasks; we provide an extension of such attacks to regression tasks such as steering angle prediction models.
    \item We validate the proposed methods using qualitative and quantitative analysis on a well-known autonomous navigation dataset to showcase its promise. We also show that a model that is adversarially trained using the images generated by our method provides significant robustness to the model.
\end{itemize}

The remainder of this paper is organized as follows. In Section \ref{Related_work}, we review previous related efforts in the areas of autonomous navigation and adversarial attacks. We describe our methodology in Section \ref{Methodology}, with a perspective of how this provides a more general notion of natural adversaries. Our implementation details, experiments and results are shown in Sections \ref{Implementation} and \ref{sec:Results} respectively, followed by conclusions and future directions in Section \ref{Conclusion}.

% \vspace{-4pt}
\section{Related Work}
\label{Related_work}
Considering the focus of this work, we present an overview of related earlier efforts from perspectives of both testing the robustness of vision models in autonomous navigation, as well as adversarial attacks in general. In each of these discussions, we present the limitations of the existing efforts and the scope of improvement when the proposed method is used.
%We divide this Section into two logical parts: in the first, we introduce classic adversarial attacks, their formulation, and physical world attacks, in the second one we dig into works which study the effect of weather on Autonomous navigation models. In each subsection, we touch upon its limitations and demonstrate the scope of improvement when our idea is used. 

% \vspace{-2pt}
\subsection{Adversarial Attacks}
\label{subsec:Adversarial_attacks_related_work}
The conventional notion of an adversarial attack is generally formalized as: given a classification model $f$, and input image $\textbf{x}$, we define a perturbation $\delta$ as a quantity that is added to $\textbf{x}$ such that:
\begin{equation}
    \arg\max f(\textbf{x} + \delta) \ne  \arg\max f(\textbf{x}) \; \textnormal{ and } ||\delta|| \leq \epsilon
\label{eq:additive_adversarial}
\end{equation} 

The $\delta$ defined above has usually been specific to the input and found through methods like FGSM \cite{fgsm}, JSMA \cite{jsma}, and the more recent PGD \cite{pgd}. The $\delta$ may also be common to the entire dataset and may be found iteratively like in the case of UAP \cite{uap} or may be generated by a GAN \cite{NAG_paper}. For the interested reader, a detailed survey of these methods is provided by Akhtar et al. in \cite{akhtar2018threat}. All these methods attempt to add a perturbation to the image, which can fool the trained model. Our proposed work is different in two ways from these efforts: (i) we extend the concept of adversaries to go beyond additive perturbations to natural perturbations such as induced by weather changes in disturbing the model (similar to \cite{hendrycks2019natural}, can be considered implicitly adversarial); (ii) most existing efforts focus on adversarial attacks in classification settings, and there has not been much effort to define an adversary in a regression-based setting. We propose an adversarial loss for regression models (steering angle prediction, in particular) in this work. %and, we also generate the adversarial sample itself from another network, hence generalizing the way adversarial attacks are defined.

\paragraph{Attacking Autonomous Navigation Models:} There have been a few explicit efforts in the recent past to develop adversarial attacks for vision models in self-driving cars. Works such as that of Eykholt et al. \cite{dawn_song_physical_attack} utilize adversarial patches \cite{adversarial_patch} which are physically added to objects like traffic lights to fool the model. Similarly, Zhang et al. \cite{CAMOU} resort to physically camouflaging cars to fool object detector models used in autonomous navigation systems, to test the robustness of the model. We observe that these efforts rely on physical changes in the environment to attack the black box models, for which human effort in case of large scale testing may be prohibitive.  On the other hand, it has been shown that weather-induced environment changes naturally result in implicit adversarial circumstances for the model involved \cite{guardian_incident}, and robustness to such weather-adversarial samples is also critical for models in autonomous navigation. We hence focus on generating natural-looking images of existing scenes affected by weather conditions (in particular, fog, in this work).

%On the one hand, we have these physically reliant adverbial attacks, while on the other we have naturally caused perturbations in the form of weather conditions like rain and Fog. Our work focuses on leveraging these weather conditions as adversaries to the model. 
%Our methodology is such that these weather conditions are generated by the network itself, hence minimizing the amount of physical effort required.
% \vspace{-2pt}
\subsection{Testing Autonomous Navigation Models with Weather Changes}
Since the advent of Generative Adversarial Networks (GANs), there have been limited efforts that have explicitly attempted to analyze the effect of different weather conditions on steering angle prediction models in self-driving cars. DeepTest \cite{deeptest} used synthetic images generated using Photoshop to study the impact of rain and fog on the predicted steering angle. The manpower required for a large-scale deployment is prohibitive for such an approach. DeepRoad \cite{deeproad} tried to automate the same using generative models instead. DeepRoad learns the translation to different weather conditions; however, the sole goal for this work is to perform image-to-image translation with no explicit goal to `fool' the model or obtain any minimum deviation of steering angle from the ground truth. In this work, we bring together adversarial and unpaired image-image translation perspectives to produce a minimum deviation in steering angle prediction in the produced images. We also show in Section \ref{sec:Results} that our method to generate fog adversaries causes an average perturbation of nearly 1 radian($\sim60 $ degrees), which can serve as a rigorous platform to test vision models in autonomous navigation. 
% \vspace{-4pt}
\subsection{Fog Generation}
Adverse weather conditions such as fog are common in day-to-day life. There have been very few efforts based on image processing and filters to generate fog in images. We note that most related work in this direction based on image processing focus on defogging \cite{defog1,defog2}, whereas our work is focused on the generation of fog. Li, et al \cite{reside} and Sakaridis et al\cite{zurich_synth_fog} attempt to generate fog using image processing methods with a strong prior. A prior is usually carefully constructed based on handcrafted features, such as texture and brightness, from the image. Unfortunately, using such priors automatically restricts the approach to constraints on expected intensity, texture, and other features present in the image. Besides, in settings of autonomous navigation, handcrafted priors do not scale to the significant variations in scenes and environments, as well as significant variations in a single scene due to effects induced by factors such as light-and-shadow and fast motion. In this work, we attempt to provide an automated method to generate fog images that are intended to distort steering angle prediction models, to improve robustness of such models. We show later in this paper that performing adversarial training using the images generated by our method provides significant robustness to the model.

%d effet and the significant variations in scenes may cause such priors to 
%For example, let us use a prior, that is capable of adding fog to bright, sunny images.  It would automatically have difficulty when it is posed with images that contain shadows. 
%We hence require models which are capable of generalizing to a wide range of input scenarios. Therefore, we rely on a Generative model-based approach to learn the generation of fog.
% \vspace{-4pt}
\section{Methodology}
\label{Methodology}
%\subsection{Generic Adversary Formulation}
When a carefully crafted perturbation is added to the image such that it causes the network to misclassify, we call it an adversarial perturbation (as in Sec \ref{subsec:Adversarial_attacks_related_work}). For a classifier network $f$ and an input image $\textbf{x}$; the perturbation, $\phi$ applied on it may be defined as: 
\begin{equation}
\begin{split}
    \arg\max \textnormal{  }f(\phi(\textbf{x})) \ne \arg\max \textnormal{  } f(\textbf{x})
\end{split}
\label{eq:adv_gen_def}
\end{equation}
To ensure visual similarity, the adversarial image should be within $\epsilon$ (very small) distance of the original input. This constraint is generally include to the above problem as:
\begin{equation}
    \textnormal{s.t } ||\phi(\textbf{x}) - \textbf{x}|| \leq \epsilon
\label{eq:visual_sim}
\end{equation}
In the case of an additive perturbation, we define $\phi$ as:
\begin{equation*}
    \phi(\textbf{x}) = \textbf{x} + \delta
\end{equation*}
We observe that this results in the same set of equations as earlier in Eqn \ref{eq:additive_adversarial}.

We now use this definition to extend the typical form of adversarial perturbations, by allowing $\phi$ to be more than simple additive perturbations. We can define $\phi$ as a multiplicative noise or a filter, or a neural network itself can model it. We expect any transformation created by $\phi$ to be valid in such a setting as long it guarantees task-perceptual similarity, discussed below. 
% \vspace{-6pt}
\paragraph{Task-Perceptual Similarity.} 
In the case of images, the adversarial attacks performed are such that both the adversarial image $\phi(\textbf{x})$ and original image $\textbf{x}$ are perceived to be similar. This usually includes visual similarity like in the case of many popular attacks like FGSM \cite{fgsm} and JSMA \cite{jsma}. 
Recent works have shown that simple image transformations like rotation, scaling, and the translation are sufficient to fool the network \cite{engstrom2017rotation}\cite{scaling_rotation_translation}. These efforts do not emphasize complete visual similarity, yet we humans find them to be perceptually similar, i.e., they are interpreted in the same way by the human visual system. Following this, we can extend the definition of adversaries beyond visual similarity to something more intuitive: \textit{task-perceptual similarity}. 

To ensure the same, in Eqn \ref{eq:adv_gen_def}, we expect the transformations created by $\phi$ to be such that the result predicted by humans for the given task is the same for both the original input and the adversary. The transformation need not necessarily guarantee visual similarity, but the task at hand needs to be perceived in the same way by the human. Examples of such transformations for images include contrast and brightness changes, blurring, rotation, scaling, sharpening, whitening, the addition of noise, etc. In all of these cases or a combination of these, humans are capable of perceiving the image similarly; this is, however, not the case with neural network models. One should note that visual similarity generally guarantees task-perceptual Similarity, but the converse need not hold. In Eqn \ref{eq:adv_gen_def}, in order to guarantee visual similarity, one may simply add a constraint like that of Eqn \ref{eq:visual_sim}.
% \vspace{-2pt}
\subsection{Adversarial Attack on Regression Models}
From the previous definition of adversarial perturbations, we may redefine it for a regression network, $N$ as: 
% \vspace{-0.5pt}
\begin{equation}
    ||N(\phi(\textbf{x})) - N(\textbf{x})|| \geq \theta \label{eq:regress_adv_eq}
\end{equation}
This implies we want the perturbation applied to the image $\textbf{x}$ to cause a minimum deviation of $\theta$. As stated earlier in Section \ref{subsec:Adversarial_attacks_related_work}, $\phi$ can be purely additive or another function captured by a neural network. We may add additional constraints on $\phi$ for visual similarity or sparsity. 
We can then simplify Eqn \ref{eq:regress_adv_eq} to:
\begin{align}
    ||N(\phi(\textbf{x})) - N(\textbf{x})|| - \theta \geq 0 \nonumber \\
    \implies \theta - ||N(\phi(\textbf{x})) - N(\textbf{x})|| \leq 0
\label{eq:regress_adv_explanation}
\end{align}
Hence, we define regression loss, $L_{regress}$ required for creating an adversarial sample for input $\textbf{x}$  as:
\begin{align}
    \min_{\phi} L_{regress} = \min_{\phi} (\theta-||N(\phi(\textbf{x})) - N(\textbf{x})||)
\label{eq:regress_adv}
\end{align}
By minimizing $L_{regress}$, we can find the perturbation, $\phi$ which ensures that a minimum deviation is caused for every input sample, $\textbf{x}$.
% \vspace{-2pt}
\subsection{Proposed Idea}
\label{Proposed Idea}
Steering angle prediction models are fundamentally designed to be regression models, and take the input scene to predict the angle of the steering wheel in the range of [$-\pi,\pi$]. 
We have previously seen in Fig \ref{fig:steering_sample} that adverse weather conditions like fog affect the steering angle predicted by a model. Fog may be represented as a mixture of blurring and whitening on an image. We also observe that a foggy image is perceived in a similar manner as the normal one by a human, especially when it comes to tasks like steering angle prediction. To elaborate further, in Fig.\ref{fig:steering_sample}, we as humans expect the same steering angle for both the sets of images but the network does not. We leverage this fact to design a testbed for steering angle models using adverse weather conditions. Hence, combining the previous two sections, we can utilize weather conditions like fog to attack steering angle prediction models adversarially.

To achieve this goal, we define $N$ as our steering angle predictor in Eqn \ref{eq:regress_adv} and $\phi$ as our adversarial weather generator for a given sample $\textbf{x}$. We train a generator, $\phi$ to learn the transformation (Sec \ref{subsec:Loss_formulation}) from normal weather to foggy. The weights of $\phi$ will be obtained by minimizing Eqn \ref{eq:regress_adv}.

We now have $\phi$, a generator being modeled by a neural network and a continuous-valued discriminator $N$. The $\theta$ in the equation naturally becomes the minimum steering angle deviation we desire.
% \vspace{-2pt}
\subsection{Loss Formulation}
\label{subsec:Loss_formulation}
We train a CycleGAN \cite{CycleGAN2017} to learn the translation between the normal (sunny) weather (\textit{Domain A}) and the adverse weather condition images (\textit{Domain B}). $\phi_{AB}$ is the Generator responsible for the translation from Domain A to B and $\phi_{BA}$ for the reverse. Similarly, $D_A$ is the Discriminator responsible for Domain A and $D_B$ for Domain B. CycleGAN utilizes adversarial losses across its generated images. It also uses a cycle-consistency loss which ensures that when an image $\textbf{x}$ belonging to Domain A, is translated from Domain A to B and back to Domain A, remains the same. i.e., 
\begin{equation*}
    \textbf{x} \approx \phi_{BA}(\phi_{AB}(\textbf{x}))
    \label{Eq:Cycle_eq_1}
\end{equation*}
This also applies in a similar fashion for images from Domain B. Hence, CycleGAN loss in total consists of: 
\footnote{For complete details of the CycleGAN loss, please refer Zhu et al\cite{CycleGAN2017}.}
\begin{align*}
    L_{CycleGAN}  = L_{cycle}(\phi_{AB}, \phi_{BA}) +  \\  L_{adversarial}(\phi_{AB}, \phi_{BA}, D_A, D_B)
\end{align*}
We augment the CycleGAN losses with the regression loss (Eqn \ref{eq:regress_adv}) and train this combined loss. 

\begin{figure*}[htbp]
\centering
  \includegraphics[width=\textwidth,height=6cm]{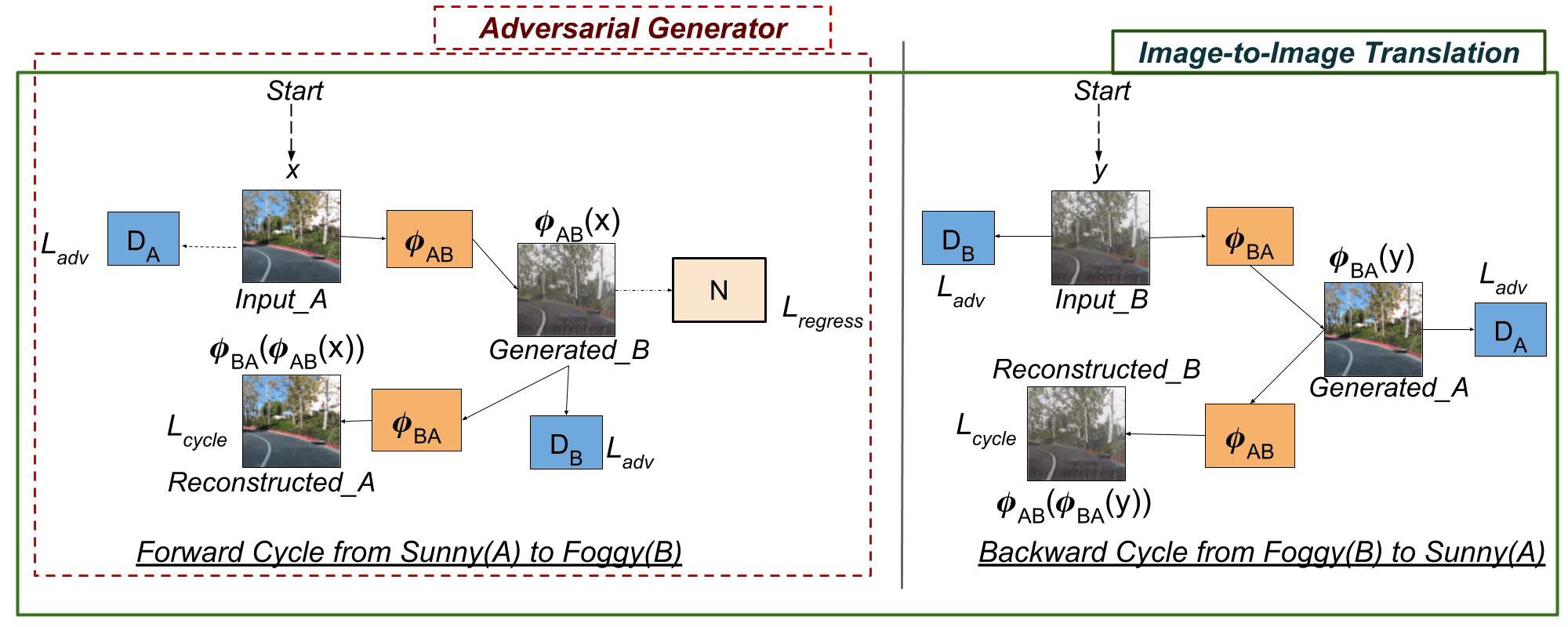}
\caption{Summary of the framework followed to train the CycleGAN network to fool the steering angle predictor $N$. The image-to image translation part of the system learns the translation between Normal, Sunny weather(Domain A) and Foggy weather(Domain B). The adversarial generator part helps $\phi_{AB}$ generate foggy images which can cause a minimum deviation($\theta$) in the predictions of $N$.}
% \vspace{-10pt}
\label{fig:method}
\end{figure*}

Hence the net loss that the CycleGAN is trained on:
\begin{align}
    L_{total} = (1-\alpha)L_{CycleGAN} + \alpha L_{regress} \\
    \textnormal{where, }L_{regress} =  \theta - ||N(\phi_{AB}(\textbf{x}))- N(\textbf{x})||
\label{Eq:CycleGAN_loss_plus_regress}
\end{align}
In the above equation $\alpha$ is a multiplier lying between $(0,1)$. 
% The $L_{CycleGAN}$ loss applies to both the generators present in the CycleGAN: $\phi_{AB},\phi_{BA} $ whereas the $L_{regress}$ applies only to Generator that converts from normal to foggy images, $\phi_{AB}$.
We summarize the working of the network in Figure \ref{fig:method} and also in Algorithm \ref{alg:train}.

% We train the combined architecture, with the Regression loss. One may notice that in this case, $\phi_{AB}$ works as a generator and the steering angle predictor works like continuous-valued  Discriminator. \\

One may note that we can use any other domain translation model instead of CycleGAN to achieve the same goal. To demonstrate the efficiency of our concept, we run our experiments on another recently popular model: DistanceGAN \cite{DistanceGAN}. This model utilizes CycleGAN losses along with a distance loss to ensure that the distance between a pair of samples is maintained across both the domains.
\vspace{-0.3cm}
\begin{algorithm}
\caption{CycleGAN training pseudo code}\label{alg:train}

\hspace*{\algorithmicindent} \textbf{Input} $X\gets$ Training samples from Domain $A$\\
\hspace*{\algorithmicindent} \textbf{Input} $Y\gets$ Training samples from Domain $B$\\
\hspace*{\algorithmicindent} \textbf{Input} $N\gets$ Steering Angle Model to be Fooled\\
\hspace*{\algorithmicindent} \textbf{Input} $T\gets$ Number of epochs to train \\
\hspace*{\algorithmicindent} \textbf{Input} $\theta\gets$ Minimum deviation desired\\
\hspace*{\algorithmicindent} \textbf{Output} $\{\phi_{AB},\phi_{BA},D_A,D_B\}$
\begin{algorithmic}[1]
\Procedure{Generate}{$ $}
\For{$t$ in $\{1...T\}$}
\State Draw $m$ training samples $\{x_1, .. x_m\}$ from X
\State Draw $m$ training samples $\{y_1, .. y_m\}$ from Y
\For{$i$ in $\{1,...m\}$}
    \State $x \gets x_i$ ; $y \gets y_i$ 
    \State Compute: $\hat{y} \gets \phi_{AB}(x)$ \Comment{Forward Cycle}
    \State Compute: $\hat{x} \gets \phi_{BA}(y)$ \Comment{Backward Cycle}
    \State Compute the losses:
    \State\hspace*{\algorithmicindent}$L_{regress}(\phi_{AB},x)$ \Comment{Eq. \ref{Eq:CycleGAN_loss_plus_regress}}
    \State\hspace*{\algorithmicindent} $L_{cycle}(\phi_{AB}, \phi_{BA}, x, y, \hat{x}, \hat{y})$
    \State Update Generators\\
     \vspace{-0.2cm}
    \State Compute Discriminator losses:
    \State $L_{adversarial}(\phi_{AB}, \phi_{BA}, D_A, D_B, x, y, \hat{x}, \hat{y})$
    \State Update Discriminators
    
\EndFor
\EndFor
% \While{$r\not=0$}\Comment{We have the answer if r is 0}
% \State $a\gets b$
% \State $b\gets r$
% \State $r\gets a\bmod b$
% \EndWhile\label{euclidendwhile}
% \State \textbf{return} $b$\Comment{The gcd is b}
\EndProcedure
\end{algorithmic}
\end{algorithm}

\section{Implementation}
\label{Implementation}

\paragraph{Network Architecture:} For the domain translation between normal to foggy weather, we use the CycleGAN model. The generators in the model use the same architecture involving Resnet-9 blocks as described in \cite{CycleGAN2017}. We also utilize similar architecture for the DistanceGAN model\cite{DistanceGAN}.\footnote{Project Webpage: \href{https://code-assasin.github.io/little_fog/}{https://code-assasin.github.io/little$\_$fog/}} 
For the steering angle models, we use AutoPilot\cite{sully_chen_autopilot} which is an improvement on NVIDIA PilotNet model\cite{nvidia_pilotnet}.  The model is adapted with an additional convolution layer for $128 \times 128$ images. We also perform similar experiments on the architecture developed by Comma AI \cite{comma}. 
% \vspace{-0.4cm}
\paragraph{Datasets used:} 
There are many weather conditions which are capable of causing a large deviation in steering angle predictions. However, due to unavailability of datasets with heavy snow or rainy weather conditions, we restrict ourselves to foggy conditions \cite{Foggy_zurich}. We, however, note that our framework is generic and can be extended to other weather conditions.

We train the CycleGAN model\cite{CycleGAN2017} to learn the domain translation from normal, sunny weather to foggy weather conditions. The sunny weather condition images are obtained from the widely used SullyChen's dataset \cite{sully_chen_autopilot}. We chose this dataset over the Udacity dataset \cite{udacity_challenge} due to its better quality. The Udacity dataset has many images with blank bright spots or hazy regions, making it difficult for the Generator to learn the translation to its equivalent foggy condition. Most other steering angle based datasets are either virtual \cite{carla} or of poor quality with minimal variations \cite{comma}, and hence not suitable for this work.

Our foggy weather images (to train the CycleGAN) are taken from the Foggy Zurich dataset \cite{Foggy_zurich}, which consists of 3.7k high-quality images collected during the occurrence of fog in and around Zurich. The steering angle prediction model is trained on the Sullychen dataset \cite{sully_chen_autopilot}, and the MSE is shown in  Table \ref{tab:Steering models}.
% \vspace{-0.4cm}
\paragraph{Preprocessing:} In order to train the CycleGAN, the normal weather images from Sully Chen's dataset \cite{sully_chen_autopilot} are sampled since many of the frames consist of similar scenes. After sampling (Systematic sampling with an offset of 12 frames), we obtain nearly 3.7k images for training the CycleGAN. The foggy images from \cite{Foggy_zurich} consist of a wiper and a dashboard in their scenes, something absent in the normal images dataset. Hence we crop the bottom few pixels of the foggy images to train the CycleGAN. Both the dataset images are resized to $128 \times 128$ pixels and then used to train the CycleGAN. For the steering angle models, however, we use the entire Sully Chen dataset \cite{sully_chen_autopilot}, split into a train and test set.  The images are resized to $128 \times 128$ and normalized between $[-1, +1]$.
% \vspace{-0.4cm}
\paragraph{Training details:} 
We train the CycleGAN model along with the regression loss to learn the translation between normal and foggy weather conditions (Eqn \ref{eq:regress_adv}). The parameters used in case of the CycleGAN losses are mostly the same as those mentioned in \cite{CycleGAN2017} except for the identity loss having a multiplier ($\lambda_{identity}$) of $3$ . The model is then trained with both losses. The value of $\alpha$ is chosen to be $0.2$ and $\theta$ is $0.5$ radians ($\sim30$ degrees).
% \vspace{-0.4cm}
\paragraph{DistanceGAN model: } We also train the DistanceGAN \cite{DistanceGAN} model with similar preprocessing, as mentioned above. We used the hyperparameters as used in the code provided by \cite{DistanceGAN}, with the exception of the identity loss. We choose the value of $\alpha$ to be $0.07$ and $\theta$ as $0.5$ radians.

% shift model to one column
\begin{table}[htbp!]
\centering
\footnotesize
\begin{tabular}{ccc}
\hline
Model & Train Error & Test Error \\
\hline
AutoPilot\cite{sully_chen_autopilot}  & 0.0185 & 0.0448 \\
% PilotNet\cite{nvidia_pilotnet} & 0.0510 & 0.0470 \\
Comma AI\cite{comma} & 0.0198 & 0.0551 \\
\hline
\end{tabular}
\caption{Test + Train error for different steering angle models}
% \vspace{-0.4cm}
\label{tab:Steering models}
\end{table}

% \vspace{-0.4cm}
\section{Results}
\label{sec:Results}
\begin{figure*}[htbp]
\centering
  \includegraphics[width=\textwidth, height=6.5cm]{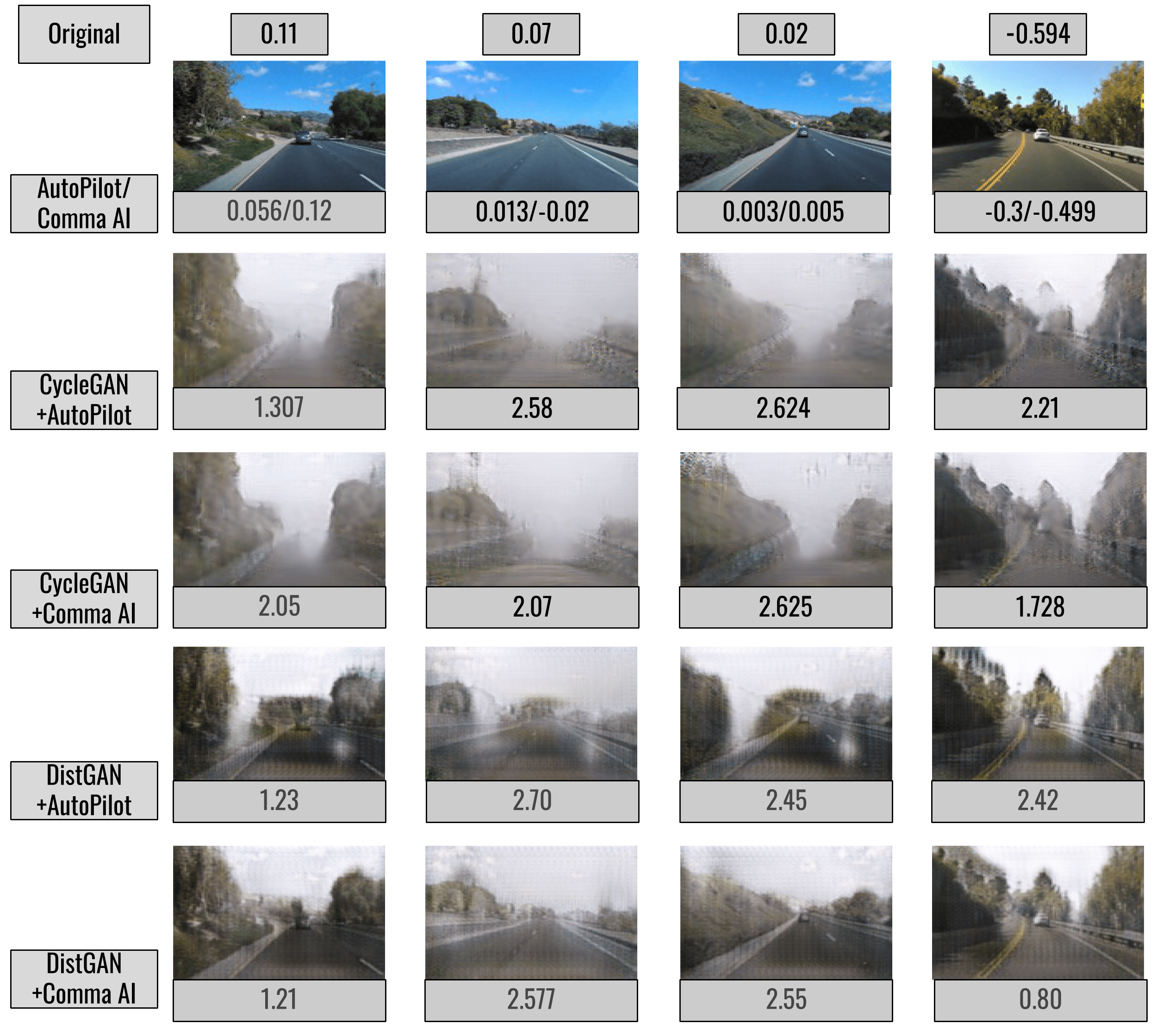}
\caption{Fooling Models: Ground truth Steering Angle (in radians) for each of the original test samples. The angles right below indicate the ordered pair of predicted steering angle by AutoPilot and Comma AI respectively. From the second row onward, we indicate the image translation model used and respective steering model it was trained on. The angle below each of those images indicates the prediction by the steering model for the generated foggy image.}
% \vspace{-4pt}
  \label{fig:Results}
\end{figure*}

% \vspace{-4pt}
We showcase the results of our trained model in Fig.\ref{fig:Results} along with the steering angle, predicted for each image. We observe that the second image has nearly a $180$ degree change in the predicted angle, indicating how dangerous these adverse weather conditions can be, further encouraging the need for testbeds provided by methods like ours. 

% \begin{figure*}[htbp]
% \centering
%   \includegraphics[width=\textwidth, height=6cm]{images/res_fig.jpg}
% \caption{Steering Angle (in radians) predicted for each image by the PilotNet Model. Row below are images generated using the proposed method. }
%   \label{fig:Results}
% \end{figure*}

\subsection{Subjective Image Quality Assessment}
% \vspace{-4pt}
We assess the realism of the foggy images generated by the CycleGAN after being trained with the Regression Loss. We asked 10 participants to assess the quality of the generated foggy images. These participants had never seen the normal, sunny version of the image before and were asked to judge the realism of the foggy image provided to them. We use the foggy images from the Foggy Zurich dataset as the control for these experiments. We ran this experiment for different variations of the CycleGAN model used. We observed that on an average (for both steering prediction models), nearly $48\%$ ($44\%$ for DistanceGAN) of the participants found the fog generated by utilizing Regression loss to be real enough compared to Foggy images from the Zurich dataset \cite{Foggy_zurich}. Additionally, we asked participants to choose the between the foggy images produced by CycleGAN (or DistanceGAN) and those produced when regression loss was added. We asked them to compare theses images based on their quality. We found that nearly $43\%$ ($48\%$ for DistanceGAN model) of the people found that CycleGAN with regression loss produces better images. 
Considering this is nearly half the participants, we find that our method produces images of comparable quality as the original foggy images. 

% The results for the same are shown in Table \ref{tab:IQA normal}. Additionally, we asked participants to choose the between the foggy images produced by CycleGAN (or DistanceGAN) alone, and those produced when regression loss was added based on their quality (Table \ref{tab:IQA with and without}).

\subsection{Objective Image Quality Assessment}
% \vspace{-4pt}
We use standard Image quality assessment metrics like MSE, PSNR, SSIM\cite{ssim} to compare the quality of the normal sunny image, w.r.t its foggy counterpart for different flavors of the models used. We report the results for the same in Table \ref{tab:IQA normal}. In addition, we compare the image quality of foggy images produced using produced by CycleGAN alone and those produced when Regression loss was added (see Table \ref{tab:IQA with and without}).

We see from Tables \ref{tab:IQA normal} and \ref{tab:IQA with and without} that the regression loss indeed causes a visible change in the image. Visibly also the quality of the images without regression loss seems to be better than those produced with it. To study the efficacy of regression loss, we compute the difference in steering angle predicted for foggy images produced by the CycleGAN model with and without regression loss, i.e., we compute:
\begin{equation*}
    ||N(\phi_{foggy}(x))- N(\hat{\phi}_{foggy}(x))|| 
\end{equation*}
$\phi_{foggy}$ is the CycleGAN Generator responsible for the translation from normal to foggy domain and $\hat{\phi}_{foggy}$ is the same generator trained with the Regression Loss. From Table \ref{tab:Deviation}, we can clearly see that the deviation produced with and without our loss is clearly very significant for both the CycleGAN and DistanceGAN model. 

\begin{table}[htbp!]
\centering
\footnotesize
\begin{tabular}{cc}
\hline
Method & Deviation Caused \\
\hline
Cycle vs Cycle+Regress (AutoPilot)   &  $1.09\pm0.9$   \\
Cycle vs Cycle+Regress (Comma AI)   &  $0.76\pm0.7$   \\
Distance vs Distance+Regress(AutoPilot)   &  $1.31\pm0.5$\\
Distance vs Distance+Regress(Comma AI)   &  $0.58\pm0.54$\\
\hline
\end{tabular}
\caption{Deviation caused with regress loss in comparison to the original models (Calculated using Eq.\ref{Eq:CycleGAN_loss_plus_regress}) }
% \vspace{-4pt}
\label{tab:Deviation}
\end{table}

%  For each write mean and variance.. 
% \begin{table}[htbp]
% \centering
% \begin{tabular}{cc}
% \hline
% Method & \% Real \\
% \hline
% \hline
% Cycle    & 46.6\\
% +Regress(AutoPilot)  &  47.5 \\
% +Regress(Comma AI)  &  47\\
% \hline
% Distance  & 44\\
% +Regress(AutoPilot) &  59  \\
% \hline
% \hline
% \end{tabular}
% \\
% \caption{Subjective IQA normal w.r.t respective foggy counterpart using different methods}
% \label{tab:IQA_subjective}
% \end{table}
% \vspace{-0.7cm}
\begin{table}[htbp]
% REDUCED inter column separation
\setlength{\tabcolsep}{3.5pt} 
\centering
\footnotesize
\begin{tabular}{cccc}
\hline
\hline
Method & MSE & PSNR & SSIM \\
\hline
\hline
Cycle  &  $3172\pm1172.4$  & $13.4\pm1.8$ & $0.51\pm0.08$\\
+Regress(AutoPilot)  &  $3024\pm1076$    & $13.7\pm1.9$   & $0.52\pm0.07$ \\
+Regress(Comma AI)  &  $3111.2\pm1086$    & $13.5\pm1.9$   & $0.52\pm0.08$\\
\hline
Distance &  $2073.2\pm663.4$    &  $15.18\pm1.4$ & $0.59\pm0.05$\\
+Regress(AutoPilot) &  $3292\pm1085.8$  & $13.2 \pm 1.56$ & $0.54\pm0.07$   \\
+Regress(Comma AI) &  $2327\pm811$  & $14.7 \pm 1.6$ & $0.61\pm0.058$   \\
\hline
\hline
\end{tabular}
\\
\caption{Objective IQA normal w.r.t respective foggy counterpart using different methods}
% \vspace{-0.4cm}
\label{tab:IQA normal}
\end{table}

\begin{figure*}[htbp!]
\centering
  \includegraphics[width=\textwidth,height=4cm]{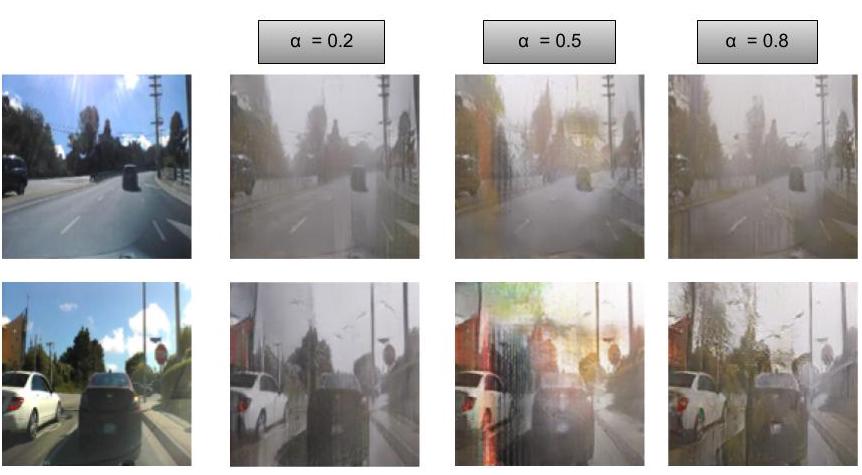}
\caption{Variation in image quality of CycleGAN for different $\alpha$ values}
\label{fig:Results_alpha}
\end{figure*}

% \begin{table*}[htbp]
% \centering
% \footnotesize
% \begin{tabular}{ccccc}
% \hline
% Method & Better Quality & MSE & PSNR & SSIM \\
% \hline
% Cycle vs Cycle+Regress(AutoPilot) &  Cycle($57\%$) &  $802.9\pm446.6$  & $19.6\pm2.04$ & $0.64\pm0.06$  \\
% Cycle vs Cycle+Regress(Comma AI) &  Cycle($57\%$) &  $868.4\pm491$  & $19.3\pm2.17$ & $0.64\pm0.07$  \\
% Distance vs Distance+Regress (AutoPilot)  &  Dist($52\%$)   &  $2276.9\pm2099$   & $16\pm3.7$  & $0.44 \pm 0.17$\\
% \hline
% \end{tabular}
% \caption{Results using different Image Quality Assessment methods}
% \label{tab:IQA with and without}
% \end{table*}

\begin{table}[htbp]
\setlength{\tabcolsep}{2pt} 
\centering
\footnotesize
\begin{tabular}{cccc}
\hline
Method & MSE & PSNR & SSIM \\
\hline
Cycle vs (AutoPilot) &  $802.9\pm446.6$  & $19.6\pm2.04$ & $0.64\pm0.06$  \\
Cycle vs (Comma AI) &  $868.4\pm491$  & $19.3\pm2.17$ & $0.64\pm0.07$  \\
Distance vs (AutoPilot) &  $2025.2\pm1253$   & $15.8\pm2.56$  & $0.52 \pm 0.07$\\
Distance vs (Comma AI) &  $1362.\pm786$   & $17.43\pm2.36$  & $0.57 \pm 0.068$\\
\hline
\end{tabular}
\caption{Results using different I.Q.A methods. For each row we compare, the model and its counterpart trained with regression loss on the respective steering model.}
\label{tab:IQA with and without}
\end{table}

\section{Ablation Studies}
\label{Ablation Studies}
We notice in Eq.\ref{Eq:CycleGAN_loss_plus_regress} that there is a subtle balance between the CycleGAN and regression loss. While $L_{CycleGAN}$ is responsible for the quality of the foggy images generated, $L_{regress}$ is responsible for causing the minimum deviation in the steering angle model. We can see that either loss overpowering the other is undesirable. Hence, a subtle balance must be created between them to achieve the end goal. To gain a deeper understanding of this, we explore variations in the hyperparameters $\alpha$ and $\theta$. For all of the experiments, we use a CycleGAN\cite{CycleGAN2017} model along with the AutoPilot\cite{sully_chen_autopilot} model as the steering angle predictor. 
\paragraph{Variation in $\alpha$.} We vary the hyper-parameter $\alpha$ (defined in Eq.\ref{Eq:CycleGAN_loss_plus_regress}) to study its effect on the images generated by the CycleGAN. We choose a constant value of $\theta = 0.5$ radians for these experiments. 
% \vspace{-3pt}
\begin{itemize}
\setlength\itemsep{-0.5em}
    \item $\alpha = 0.2:$ With a decent value of $\alpha$, as seen in Section \ref{sec:Results}, it produces fairly good results. The image quality of the CycleGAN remains pretty clear while causing the minimum required deviation.
    \item $\alpha = 0.5:$ With a moderate value of $\alpha$, although we expect to average results, the image quality of the CycleGAN is completely wrecked. 
    \item $\alpha = 0.8:$ With very high value of $\alpha$, $L_{regress}$ is given more importance. This causes the output foggy images to have lower-quality than those obtained in Section \ref{sec:Results}.
\end{itemize}
\vspace{-0.2cm}
While for $\alpha = 0.2$ it only takes 150 epochs to reach the desired goal, for higher values of $\alpha$ it seems to take more epochs to converge. For $\alpha = 0.5$ it took nearly 200 additional epochs to settle while the losses kept oscillating for $0.8$. We can also see the variation in their quality in Fig \ref{fig:Results_alpha}.

\paragraph{Variation in $\theta$.} We vary the minimum deviation $\theta$ in the Eq. \ref{Eq:CycleGAN_loss_plus_regress} to see its effect on the generated images. For all experiments with changes in $\theta$, we fix our $\alpha$ value to $0.2$.
\vspace{-3pt}
\begin{itemize}
\setlength\itemsep{-0.5em}
    \item $\theta = 0$ radians : In this case, we would ideally expect the $L_{regress}$ to be dominated by $L_{CycleGAN}$ but we observe that in a matter of few epochs the $L_{regress}$ values start becoming highly negative hence the net loss function start shifting its attention towards minimizing the regression loss without paying heed to the CycleGAN losses. This causes a lot of deterioration in the image quality of the foggy images. 

    \item $\theta = 0.5$ radians : As seen in Section \ref{sec:Results}, this value of $\theta$ relayed good results both in terms of image quality and deviation produced.
    \item $\theta = 1$ radians:  With a higher value of $\theta$ the number of epochs taken for convergence was much higher with nearly similar quality as those obtained with $\theta = 0.5$. Hence, we prefer to reduce our costs  by utilizing $\theta = 0.5$. In addition, $\theta = 0.5 $ itself is causing an average deviation of $1$ radian (Table \ref{tab:Deviation}) hence reaping the benefits without additional costs.
\end{itemize}

Similar trends, for $\theta$ and $\alpha$, are seen using the DistanceGAN \cite{DistanceGAN} model.

\begin{figure}[ht]
\centering
\subfloat[]{
    \centering
  % include first image
  \includegraphics[width=.35\textwidth,height=2cm]{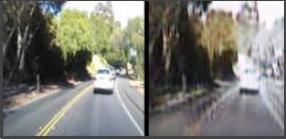}  
  \label{fig:back_regress_1}
}\\[-10pt] 
\subfloat[]{
  \centering
  % include second image
  \includegraphics[width=.35\textwidth,height=2cm]{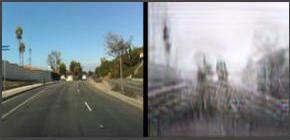}  
  \label{fig:back_regress_2}
}
\caption{ Original images (left) and foggy counterparts (right) generated using regression loss in both directions}
\vspace{-4pt}
\label{fig:back_regress}
\end{figure}

\paragraph{Regression loss on Backward Generator $\phi_{BA}$:}
From Eq. \ref{Eq:CycleGAN_loss_plus_regress} we can observe that the Regression loss is applied only to the forward cycle of the CycleGAN. Our goal is only to create adverse weather conditioned images and not vice-versa. 
Hence the Regression loss is only applied on $\phi_{AB}$. To test if adding a regression loss on backward cycle might help, we add the following term to Eq. \ref{Eq:CycleGAN_loss_plus_regress}: 
\begin{align*}
      L_{Bregress} = \theta - ||N(\phi_{BA}(y)) - N(y)||
\end{align*}
Where $y$ belongs to Domain B(foggy weather). This loss is then multiplied with $\alpha$, same as in Eq. \ref{Eq:CycleGAN_loss_plus_regress}. We train the CycleGAN model in conjunction with these losses. 
As seen in Fig. \ref{fig:back_regress}, the results of this model seems to be distorted, and of poor quality. We also observed that the loss seemed to oscillate. We believe that this occurs because backward regression loss forces the normal images generated to cause distortion. There is a good chance (as in Sec \ref{Ablation Studies}) that the regression loss dominates over the CycleGAN loss and hence, might even generate blank images which are perfectly capable of causing the desired deviation in the steering angle. To prevent this from happening, we ensure that the generated image is perfectly capable of being returned to the original (Eq. \ref{Eq:Cycle_eq_1}). We hence do not include the adversarial loss term in $\phi_{BA}$.

% \begin{figure*}[htbp!]
% \centering
%  \includegraphics[height=3cm,width=0.75\textwidth]{images/back_and_front_regress.png}
% \caption{Distorted Images of poor quality seen using regression loss on backward cycle}
% \label{fig:back_and_front_regress}
% \end{figure*}

\vspace{-0.4cm}
\paragraph{Defending against Fog:}
We train the AutoPilot and Comma AI models with the foggy counterparts of the training dataset \cite{sully_chen_autopilot}, generated using the proposed method. We then test the model on the foggy counterpart of the test set. From Table \ref{tab:Deviation_defense}, there is a clear improvement in the deviation caused once the model is adversarially trained using the foggy images generated by our method. 

\begin{table}[htbp!]
\centering
\footnotesize
\begin{tabular}{ccc}
\hline
\hline
Method & Trained on & Deviation Caused \\
\hline
\hline
AutoPilot  &  Normal & $1.81\pm1.03$   \\
AutoPilot  &  Foggy & $0.17\pm0.4$   \\
\hline
Comma AI  &  Normal & $1.88\pm0.84$   \\
Comma AI  &  Foggy & $0.2\pm0.7$   \\
% Distance vs Distance+Regress   &  $1.8\pm1.25$\\
\hline
\hline
\end{tabular}
\caption{Deviation caused with regress loss in comparison to the original models}
\vspace{-4pt}
\label{tab:Deviation_defense}
\end{table}

\section{Conclusions}
\label{Conclusion}
\vspace{-4pt}
As part of our work, we introduced a more generic definition of adversarial perturbations. Our definition makes use of the more understandable Perceptual Similarity rather than Visual Similarity. We have also introduced a manner in which adversarial perturbations may be used to fool regression-based networks such that they cause a minimum deviation. We then showed how this could be applied to steering angle models to generate sufficient Fog to produce a minimum deviation in the scene. In the future, with the help of better datasets, we would like to construct more stringent testbeds to evaluate the credibility of autonomous navigation.
% Add we show how to defend also...

% \section{Future Work}
% \label{sec:Future_work}
% \begin{itemize}
%     % \item \textit{Not valid for dense Fog, since it defies perceptual Similarity. That being said, it would be nice to introduce control over the amount of fog produced in an image. This would help generate various types of test cases for the same scene.}
%     \item \textit{Only formulated for feed-forward steering models. Right now given how cycle gan architecture works. In the future would like a universal model for all types.. }
% \end{itemize}

{\small
\bibliographystyle{ieee}
\bibliography{egbib}

\begin{thebibliography}{1}\itemsep=-1pt

\bibitem{Alpher02}
A.~Alpher.
\newblock Frobnication.
\newblock {\em Journal of Foo}, 12(1):234--778, 2002.

\bibitem{Alpher03}
A.~Alpher and J.~P.~N. Fotheringham-Smythe.
\newblock Frobnication revisited.
\newblock {\em Journal of Foo}, 13(1):234--778, 2003.

\bibitem{Alpher04}
A.~Alpher, J.~P.~N. Fotheringham-Smythe, and G.~Gamow.
\newblock Can a machine frobnicate?
\newblock {\em Journal of Foo}, 14(1):234--778, 2004.

\bibitem{Authors06b}
Authors.
\newblock Frobnication tutorial, 2006.
\newblock Supplied as additional material {\tt tr.pdf}.

\bibitem{Authors06}
Authors.
\newblock The frobnicatable foo filter, 2011.
\newblock Face and Gesture submission ID 324. Supplied as additional material
  {\tt fg324.pdf}.

\end{thebibliography}


\begin{thebibliography}{10}\itemsep=-1pt

\bibitem{weathernewsarticle2014}
E.~Ackerman.
\newblock {\em Korean Competition Shows Weather Still a Challenge for
  Autonomous Cars}, 2014 (accessed July 25, 2019).

\bibitem{akhtar2018threat}
N.~Akhtar and A.~Mian.
\newblock Threat of adversarial attacks on deep learning in computer vision: A
  survey.
\newblock {\em IEEE Access}, 6:14410--14430, 2018.

\bibitem{scaling_rotation_translation}
A.~Azulay and Y.~Weiss.
\newblock Why do deep convolutional networks generalize so poorly to small
  image transformations?
\newblock {\em arXiv preprint arXiv:1805.12177}, 2018.

\bibitem{DistanceGAN}
S.~Benaim and L.~Wolf.
\newblock One-sided unsupervised domain mapping.
\newblock In {\em NIPS}, 2017.

\bibitem{nvidia_pilotnet}
M.~Bojarski, D.~Del~Testa, D.~Dworakowski, B.~Firner, B.~Flepp, P.~Goyal, L.~D.
  Jackel, M.~Monfort, U.~Muller, J.~Zhang, et~al.
\newblock End to end learning for self-driving cars.
\newblock {\em arXiv preprint arXiv:1604.07316}, 2016.

\bibitem{adversarial_patch}
T.~B. Brown, D.~Man{\'e}, A.~Roy, M.~Abadi, and J.~Gilmer.
\newblock Adversarial patch.
\newblock {\em arXiv preprint arXiv:1712.09665}, 2017.

\bibitem{carlini_speech_attack}
N.~Carlini and D.~Wagner.
\newblock Audio adversarial examples: Targeted attacks on speech-to-text.
\newblock In {\em 2018 IEEE Security and Privacy Workshops (SPW)}, pages 1--7.
  IEEE, 2018.

\bibitem{sully_chen_autopilot}
S.~Chen.
\newblock Autopilot-tensorflow, 2016.
\newblock {\em URL https://github. com/SullyChen/Autopilot-TensorFlow}, 2016.

\bibitem{agv_attack_paper}
S.-T. Chen, C.~Cornelius, J.~Martin, and D.~H.~P. Chau.
\newblock Shapeshifter: Robust physical adversarial attack on faster r-cnn
  object detector.
\newblock In {\em Joint European Conference on Machine Learning and Knowledge
  Discovery in Databases}, pages 52--68. Springer, 2018.

\bibitem{Cordts2016Cityscapes}
M.~Cordts, M.~Omran, S.~Ramos, T.~Rehfeld, M.~Enzweiler, R.~Benenson,
  U.~Franke, S.~Roth, and B.~Schiele.
\newblock The cityscapes dataset for semantic urban scene understanding.
\newblock In {\em Proc. of the IEEE Conference on Computer Vision and Pattern
  Recognition (CVPR)}, 2016.

\bibitem{carla}
A.~Dosovitskiy, G.~Ros, F.~Codevilla, A.~Lopez, and V.~Koltun.
\newblock Carla: An open urban driving simulator.
\newblock {\em arXiv preprint arXiv:1711.03938}, 2017.

\bibitem{engstrom2017rotation}
L.~Engstrom, B.~Tran, D.~Tsipras, L.~Schmidt, and A.~Madry.
\newblock A rotation and a translation suffice: Fooling cnns with simple
  transformations.
\newblock {\em arXiv preprint arXiv:1712.02779}, 2017.

\bibitem{dawn_song_physical_attack}
I.~Evtimov, K.~Eykholt, E.~Fernandes, T.~Kohno, B.~Li, A.~Prakash, A.~Rahmati,
  and D.~Song.
\newblock Robust physical-world attacks on deep learning models.
\newblock {\em arXiv preprint arXiv:1707.08945}, 1:1, 2017.

\bibitem{GARIBOTTO199751}
G.~Garibotto, P.~Bassino, M.~Ilic, and S.~Masciangelo.
\newblock C.1 - computer vision for autonomous navigation: From research to
  applications.
\newblock In V.~Cappellini, editor, {\em Time-Varying Image Processing and
  Moving Object Recognition, 4}, pages 51 -- 56. Elsevier Science B.V.,
  Amsterdam, 1997.

\bibitem{fgsm}
I.~J. Goodfellow, J.~Shlens, and C.~Szegedy.
\newblock Explaining and harnessing adversarial examples.
\newblock {\em arXiv preprint arXiv:1412.6572}, 2014.

\bibitem{mask_rcnn}
K.~He, G.~Gkioxari, P.~Doll{\'{a}}r, and R.~B. Girshick.
\newblock Mask {R-CNN}.
\newblock {\em CoRR}, abs/1703.06870, 2017.

\bibitem{Hebert1988}
M.~Hebert.
\newblock Computer vision for autonomous navigation.
\newblock Technical report, Carnegie Mellon University, June 1988.

\bibitem{hendrycks2019natural}
D.~Hendrycks, K.~Zhao, S.~Basart, J.~Steinhardt, and D.~Song.
\newblock Natural adversarial examples, 2019.

\bibitem{JanaiGBG17}
J.~Janai, F.~G{\"{u}}ney, A.~Behl, and A.~Geiger.
\newblock Computer vision for autonomous vehicles: Problems, datasets and
  state-of-the-art.
\newblock {\em CoRR}, abs/1704.05519, 2017.

\bibitem{goodfellow_physical_world}
A.~Kurakin, I.~J. Goodfellow, and S.~Bengio.
\newblock Adversarial examples in the physical world.
\newblock {\em CoRR}, abs/1607.02533, 2016.

\bibitem{reside}
B.~Li, W.~Ren, D.~Fu, D.~Tao, D.~Feng, W.~Zeng, and Z.~Wang.
\newblock Benchmarking single-image dehazing and beyond.
\newblock {\em IEEE Transactions on Image Processing}, 28(1):492--505, 2018.

\bibitem{defog1}
Y.~Li, S.~You, M.~S. Brown, and R.~T. Tan.
\newblock Haze visibility enhancement: A survey and quantitative benchmarking.
\newblock {\em Computer Vision and Image Understanding}, 165:1--16, 2017.

\bibitem{RobotCarDatasetIJRR}
W.~Maddern, G.~Pascoe, C.~Linegar, and P.~Newman.
\newblock {1 Year, 1000km: The Oxford RobotCar Dataset}.
\newblock {\em The International Journal of Robotics Research (IJRR)},
  36(1):3--15, 2017.

\bibitem{pgd}
A.~Madry, A.~Makelov, L.~Schmidt, D.~Tsipras, and A.~Vladu.
\newblock Towards deep learning models resistant to adversarial attacks.
\newblock {\em arXiv preprint arXiv:1706.06083}, 2017.

\bibitem{instagram_2018}
D.~Mahajan, R.~Girshick, V.~Ramanathan, K.~He, M.~Paluri, Y.~Li, A.~Bharambe,
  and L.~van~der Maaten.
\newblock Exploring the limits of weakly supervised pretraining.
\newblock In {\em Proceedings of the European Conference on Computer Vision
  (ECCV)}, pages 181--196, 2018.

\bibitem{uap}
S.-M. Moosavi-Dezfooli, A.~Fawzi, O.~Fawzi, and P.~Frossard.
\newblock Universal adversarial perturbations.
\newblock In {\em Proceedings of the IEEE Conference on Computer Vision and
  Pattern Recognition}, pages 1765--1773, 2017.

\bibitem{jsma}
N.~Papernot, P.~McDaniel, S.~Jha, M.~Fredrikson, Z.~B. Celik, and A.~Swami.
\newblock The limitations of deep learning in adversarial settings.
\newblock In {\em 2016 IEEE European Symposium on Security and Privacy
  (EuroS\&P)}, pages 372--387. IEEE, 2016.

\bibitem{NAG_paper}
K.~Reddy~Mopuri, U.~Ojha, U.~Garg, and R.~Venkatesh~Babu.
\newblock Nag: Network for adversary generation.
\newblock In {\em Proceedings of the IEEE Conference on Computer Vision and
  Pattern Recognition}, pages 742--751, 2018.

\bibitem{guardian_incident}
G.~Report.
\newblock Korean competition shows weather still a challenge for autonomous
  cars, 2016.
\newblock
  https://spectrum.ieee.org/cars-that-think/transportation/advanced-cars/japan-competition-shows-weather-still-a-challenge-for-autonomous-cars.

\bibitem{synthiaCVPR2016}
G.~Ros, L.~Sellart, J.~Materzynska, D.~Vazquez, and A.~M. Lopez.
\newblock The synthia dataset: A large collection of synthetic images for
  semantic segmentation of urban scenes.
\newblock In {\em Proceedings of the IEEE conference on computer vision and
  pattern recognition}, pages 3234--3243, 2016.

\bibitem{Foggy_zurich}
C.~Sakaridis, D.~Dai, S.~Hecker, and L.~Van~Gool.
\newblock Model adaptation with synthetic and real data for semantic dense
  foggy scene understanding.
\newblock In {\em European Conference on Computer Vision (ECCV)}, pages
  707--724, 2018.

\bibitem{zurich_synth_fog}
C.~Sakaridis, D.~Dai, and L.~Van~Gool.
\newblock Semantic foggy scene understanding with synthetic data.
\newblock {\em International Journal of Computer Vision}, 126(9):973--992,
  2018.

\bibitem{comma}
E.~Santana and G.~Hotz.
\newblock Learning a driving simulator.
\newblock {\em arXiv preprint arXiv:1608.01230}, 2016.

\bibitem{face_recog_attack}
M.~Sharif, S.~Bhagavatula, L.~Bauer, and M.~K. Reiter.
\newblock Accessorize to a crime: Real and stealthy attacks on state-of-the-art
  face recognition.
\newblock In {\em Proceedings of the 2016 ACM SIGSAC Conference on Computer and
  Communications Security}, pages 1528--1540. ACM, 2016.

\bibitem{deeptest}
Y.~Tian, K.~Pei, S.~Jana, and B.~Ray.
\newblock Deeptest: Automated testing of deep-neural-network-driven autonomous
  cars.
\newblock In {\em Proceedings of the 40th international conference on software
  engineering}, pages 303--314. ACM, 2018.

\bibitem{udacity_challenge}
Udacity, 2016.
\newblock https://github.com/udacity/self-driving-car.

\bibitem{varma2019idd}
G.~Varma, A.~Subramanian, A.~Namboodiri, M.~Chandraker, and C.~Jawahar.
\newblock Idd: A dataset for exploring problems of autonomous navigation in
  unconstrained environments.
\newblock In {\em 2019 IEEE Winter Conference on Applications of Computer
  Vision (WACV)}, pages 1743--1751. IEEE, 2019.

\bibitem{ssim}
Z.~Wang, A.~C. Bovik, H.~R. Sheikh, E.~P. Simoncelli, et~al.
\newblock Image quality assessment: from error visibility to structural
  similarity.
\newblock {\em IEEE transactions on image processing}, 13(4):600--612, 2004.

\bibitem{defog2}
Y.~Xu, J.~Wen, L.~Fei, and Z.~Zhang.
\newblock Review of video and image defogging algorithms and related studies on
  image restoration and enhancement.
\newblock {\em IEEE Access}, 4:165--188, 2016.

\bibitem{deeproad}
M.~Zhang, Y.~Zhang, L.~Zhang, C.~Liu, and S.~Khurshid.
\newblock Deeproad: Gan-based metamorphic autonomous driving system testing.
\newblock {\em arXiv preprint arXiv:1802.02295}, 2018.

\bibitem{CAMOU}
Y.~Zhang, H.~Foroosh, P.~David, and B.~Gong.
\newblock {CAMOU}: Learning physical vehicle camouflages to adversarially
  attack detectors in the wild.
\newblock In {\em International Conference on Learning Representations}, 2019.

\bibitem{CycleGAN2017}
J.-Y. Zhu, T.~Park, P.~Isola, and A.~A. Efros.
\newblock Unpaired image-to-image translation using cycle-consistent
  adversarial networks.
\newblock In {\em Computer Vision (ICCV), 2017 IEEE International Conference
  on}, 2017.

\end{thebibliography}
}

\end{document}